# A Multisite, Report-Based, Centralized Infrastructure for Feedback and Monitoring of Radiology AI/ML Development and Clinical Deployment

Menashe Benjamin[1,*], Guy Engelhard[1], Alex Aisen[2], Yinon Aradi[1], Elad Benjamin[2]

[1] Philips Algotec, a Philips Co.
[2] Royal Philips
[*] Corresponding Author: menashe.benjamin@philips.com
{guy.engelhard, alex.aisen, yinon.aradi, elad.benjamin}@philips.com

## ABSTRACT

An infrastructure for multisite, geographically-distributed creation and collection of diverse, high-quality, curated and labeled radiology image data is crucial for the successful automated development, deployment, monitoring and continuous improvement of Artificial Intelligence (AI)/Machine Learning (ML) solutions in the real world. An interactive radiology reporting approach that integrates image viewing, dictation, natural language processing (NLP) and creation of hyperlinks between image findings and the report, provides localized labels during routine interpretation. These images and labels can be captured and centralized in a cloud-based system. This method provides a practical and efficient mechanism with which to monitor algorithm performance. It also supplies feedback for iterative development and quality improvement of new and existing algorithmic models. Both feedback and monitoring are achieved without burdening the radiologist. The method addresses proposed regulatory requirements for post-marketing surveillance and external data. Comprehensive multi-site data collection assists in reducing bias. Resource requirements are greatly reduced compared to dedicated retrospective expert labeling.

## HIGHLIGHTS

- AI/ML algorithms require feedback and monitoring from the field, for initial development, ongoing improvement and surveillance.
- Traditional retrospective expert review is expensive to use for feedback and doesn't address monitoring.
- Interactive radiology reporting with image location hyperlinks gathered from multiple sites and centralized in the cloud provides data for feedback and monitoring.

## 1. INTRODUCTION

The successful deployment of Artificial Intelligence (AI) and Machine learning (ML) in radiology presents many challenges [1-3]. Development of effective algorithms requires enormous amounts of data for training, testing and validation [4-6]. The best data is comprised of volumes of imaging studies collected from diverse sites, modality vendors, protocols and patient populations, with findings labeled by experienced radiologists. Labels are most useful when they identify locations on images. The strength of a label depends on many factors that affect its quality [7]. Such data is difficult to obtain on a large scale [8, 9]. Even in sophisticated facilities, it may be difficult to define, obtain and manage, especially when more than one institution is involved [10].

Once an algorithm is developed, tested, approved and deployed, regulators and good clinical practice require an ongoing process, preferably automated, for monitoring algorithm performance, such as by assessing whether radiologists agree with its output.

Labels for ML data sets have traditionally been obtained by retrospective expert review [9, 11]. Considerable effort is required to curate multi-system, multi-institutional, multi-national image collections and to perform retrospective expert labeling [12]. There is a trade-off between the effort it takes to label data, and the quality of the labels. Weakly labeled data may undermine learning performance [13], though in some circumstances even weak labels, when available in large numbers, may be sufficient [14]. A mixture of trusted and noisy labels may be useful [15].

For algorithms to be generalizable for deployment in diverse settings, performance should be verified using appropriate external data, beyond the data sets on which they are initially developed [16, 17]. Many published descriptions of algorithms lack such external validation [18]. Algorithms may fail when implemented in novel environments or presented with new data [19]. Generalizability is addressed by use of appropriate training data accounting for many factors, including disease prevalence and severity, technical modality considerations such as image resolution, varying acquisition protocols, modality vendor [20-22], and avoidance of ethnic, gender, and socioeconomic biases in the training data [5, 23-25]. Sample size, choice of validation strategy and appropriate metrics for performance evaluation are important [26].

To facilitate acquisition of the necessary training data, and to perform ongoing algorithm monitoring, appropriate tools must be integrated into the clinical workflow, allowing for practical and painless collection of radiologist feedback [27]. Algorithms must be evaluated in the context of their intended use by health care professionals, even to the extent of assessing their effect on patient outcomes when practical [17]. Yet large scale clinical trials for such evaluations may not be feasible or affordable, suggesting the need for pragmatic approaches such as assessment of performance in the field, recognizing that observational data is less robust than randomized data.

Current approaches, including retrospective expert annotation of limited datasets, are not integrated into the routine clinical workflow. They fail to provide a scalable mechanism for obtaining and recording feedback from radiologists. They also cannot monitor deployed algorithms to detect and recover from degraded performance and automate quality improvement.

To address these limitations, an infrastructure is needed for geographically-distributed creation, collection, and use of high-quality curated and labeled image data for developing and improving AI algorithms. Since radiologists will not use tools that reduce their productivity, an efficient approach to providing the necessary labels is to take advantage of their findings while studies are being interpreted, using reporting software that is seamless to their workflow.

## 1.1. The Cyclic Nature of AI Development: Feedback and Monitoring

AI/ML algorithm development is a cyclic process, in which algorithms can be developed, tested and improved in an iterative manner, and if sufficiently mature, redeployed in production, either continuously or at intervals (Fig. 1). We distinguish two steps in the process: *Feedback* and *Monitoring*. Ideally, both would be achieved by a method seamlessly integrated into the clinical workflow [28].

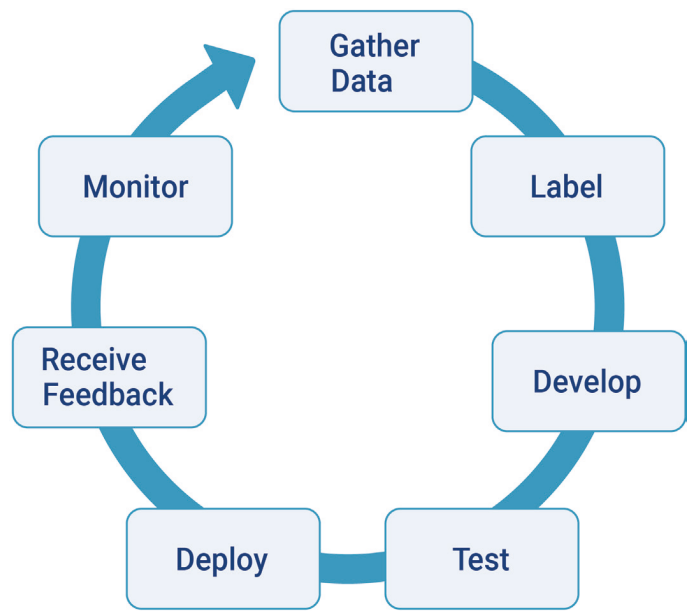

**Fig. 1.** The AI Development and Deployment Process – a continuous, never ending cycle.
Until regulatory approval, algorithms are developed, tested and improved iteratively in the background using a feedback process. After regulatory approval, they are monitored for accuracy during routine reading against labels collected from interactive radiology reports.

*Feedback* [29] is the gathering of information about algorithm performance, including the output of the algorithm as run, and the truth with which to compare it. Such feedback may be the findings of an expert, a radiologist for instance, obtained prospectively (perhaps during clinical operation) or retrospectively. It may be collected with or without exposing the expert to the algorithm output. Feedback based on reliable evidence of truth can be used for the labeling of data for training and testing of algorithms. It can be obtained in distinct testing phases, or in an operational deployment.

***Monitoring*** is the continuous gathering of feedback during operation to determine if a deployed algorithm is performing as expected, e.g., consistently with the radiologist. Safety and efficacy can be continuously assured by monitoring. Monitoring will expose deterioration in performance, including internal drift (changes in model behavior) and external drift (changes in data characteristics) [30]. When deviance is detected, alerts can be raised, and provided to the user, management and the algorithm developer. An unexpected influx of patients with new characteristics, the presence of new diseases, or the deployment of new imaging equipment or techniques, can all lead to performance degradation, which in severe cases may be catastrophic. Monitoring, tightly integrated with routine clinical reporting, may be able to detect this earlier than other approaches. Detection of performance degradation at one site may serve to alert other sites. Monitoring also safeguards against malicious attacks.

Regulatory approval of AI/ML devices, whether they be locked successive releases, or continuously learning systems, requires greater emphasis on post-marketing surveillance for continued assurance of safety and efficacy, a so-called Total Product Lifecycle regulatory approach [31, 32]. Monitoring and feedback support surveillance. Feedback and monitoring information derived from clinical reports may satisfy regulators' requirements for Real-World Data [33].

The cyclic process of improvement based on feedback must be rigorously documented, as a best practice and for regulatory compliance. Improvements must be reproducible, demonstrated and quantifiable, and serve to enhance safety and the AI value proposition. Continuous and significant improvement, as well as monitoring for performance degradation, may also serve to mitigate the phenomenon of AI "disillusionment" with underperforming algorithms [34], as well as engage the radiologists in their recognized role as information specialists [35]. Monitoring may mitigate issues similar to those evident with the early deployment of mammography Computer Aided Detection (CAD), wherein some studies suggested real world benefits did not meet initial expectations [15, 36]. The difficulty of gathering data for feedback and monitoring has been emphasized [37].

## 2.    APPROACH

### 2.1.   Leveraging the Radiology Report

The radiology report is a rich source of information [38] and may be a useful source of labels [9]. Retrospective natural language processing (NLP) can be applied to the routine clinical radiology report to obtain labels to train deep learning–based image classification models [39]. Such labels are limited and considered inferior to labels generated by explicit expert review and/or follow-up, but are easier and less expensive to obtain [40]. Concerns have been raised about the reliability of report-extracted labels, emphasizing the importance of the knowledge extraction algorithm and choice of labels [41, 42]. Quantitative approaches to measuring the value of labels have been proposed [43].

Radiology reporting solutions may incorporate features such as the inclusion of multimedia, for example key or annotated images, and interactive functionality such as hyperlinks between report text and specific findings on images (Fig. 2).

We propose that NLP integrated into the reporting process, when coupled with hyperlinks to images and locations on images, can produce stronger labels than NLP alone, in the sense that one can be more confident they represent ground truth. Such labels are approaching the gold standard of dedicated retrospective expert review, with reduced effort and less impact on efficiency.

Localization of findings within images is important. Hyperlinks are often routinely inserted during reporting for use in linking prior studies or facilitating communication to the referring practitioner, when a suitable reporting system is used [44]. These have the significant benefit of connecting findings described in report text with images. Their use often requires little additional effort.  In contrast, in the retrospective expert review scenario, precise localization may be "far too time-consuming, inaccurate, and not reproducible" [12]. However, localization information need not always be precise to be useful. Approximate information as from pointers or bounding boxes [45], typical of what radiologists use in routine clinical work, may be sufficient. The feasibility of extracting bounding boxes for AI/ML from annotations made routinely in the workflow has been demonstrated [46].

Capturing hyperlinks does not significantly disrupt the process of dictating a report, or distract the radiologist from looking at images, since a voice command may be used to insert the hyperlink, a very minimal incremental burden (Fig. 2). Radiologists create hyperlinks and reuse them when reporting new studies, even in the absence of evidence of use by referring providers [47]. Hyperlinks enable the automated generation of a multimedia report with pictures of key images, which may provide a competitive advantage if preferred by referrers [44]. Multimedia reports are increasingly desired throughout the enterprise [48]. Even in settings requiring a significant investment of radiologist or preprocessing staff effort, such as to perform tumor quantification for treatment monitoring, efficiency can be achieved using dedicated reporting software. The localized measurement annotations are also re-used for AI/ML feedback and monitoring [49, 50].

Our proposed approach leverages hyperlinks produced in routine clinical reporting to extract information that can be reused as a source of feedback and data for monitoring. This is most efficient if the reporting system is tightly integrated with the PACS workstation.

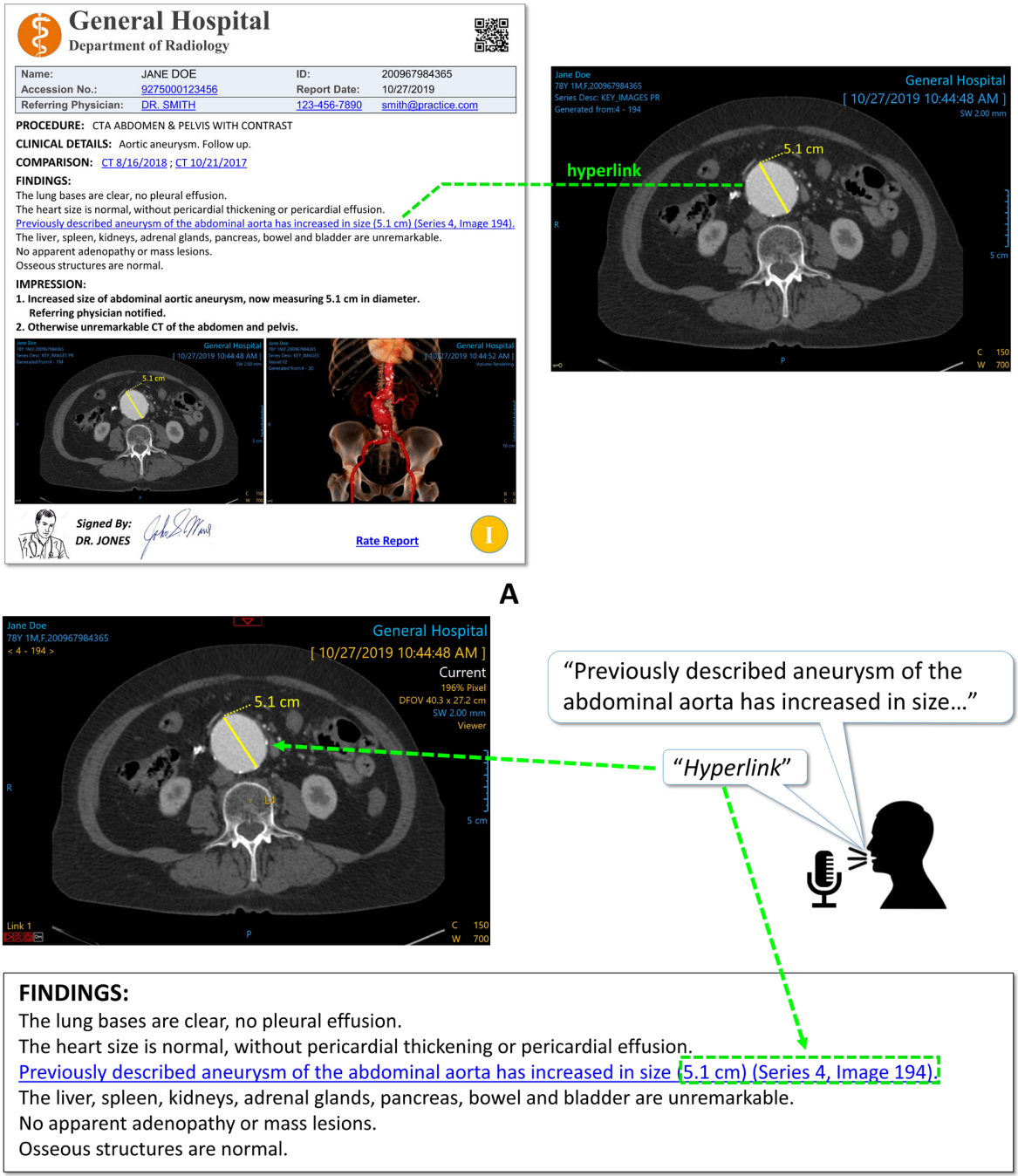

**Fig. 2.** The Interactive Report as a means of creating reliable labeled data by radiologists during routine reporting (a similar approach can be used for reporting in other specialties).

**A.** The Interactive Report contains embedded hyperlinks to locations and measurements in images, enabling the efficient generation of labeled data.

**B.** Creating hyperlinks is straightforward and easily accomplished during routine interpretation. The mechanism displayed in the illustration employs a single voice command ("Hyperlink") to associate the text to the aneurysm location and measurement.

## 2.2. Centralized Cloud Implementation

How can labels created at a site during the course of routine clinical radiology workflow at multiple sites using the interactive reporting approach be used for feedback for model updates and performance monitoring (post-marketing surveillance)?

We propose that the traditional approach of centralizing the image and label data, which has been well-established both for research and registry applications, as well as for clinical data sharing, is practical and appropriate. Though patient consent, study de-identification, ownership and business entity relationship issues are challenging to solve, they are well understood.

Incorporation of centralized image and label data acquisition for AI/ML feedback and performance monitoring into the routine production systems obviates the need for each site to build dedicated infrastructures [10, 51].

When PACS with integrated images and reporting shared across related sites is implemented (i.e., all sites have access to each other's data for clinical purposes), the centralization of images and labels is straightforward. With common access to imaging studies and supporting data, and a consistent reporting interface for all sites, even a radiology group that provides professional services to unrelated facilities can provide centralized labels [52]. Centralized data collection for feedback and monitoring will require de-identification of images and labels. The requirements for de-identification are well documented [53] and routinely applied in a manner that preserves their utility for algorithm improvement [54-56].

It is appropriate to use a cloud-based infrastructure for data collection and aggregation [57, 58]. Each source site can de-identify the images and accompanying label data before transmission to the cloud. Such a big data platform will allow multicenter research collaboration and support very large numbers of curated imaging studies, though from routine clinically deployed systems, in a similar manner to systems that have previously been used for research and development collaborations [59].

The result is a network of interconnected systems, in which the integrated PACS and reporting system at each site acts as a source of both images and labels (Fig. 3). This allows for centralized performance monitoring, and feedback for updating algorithms. The updated algorithms are then redistributed to the sites, whether deployed on-premise or as a cloud service, continuously or with successively approved releases.

Important functions of the centralized cloud component will include not only gathering of the data and model retraining, but also appropriate reporting of performance to sites, managers, algorithm developers and regulators, as well as maintenance of an audit trail of changes to software including neural network model weights.

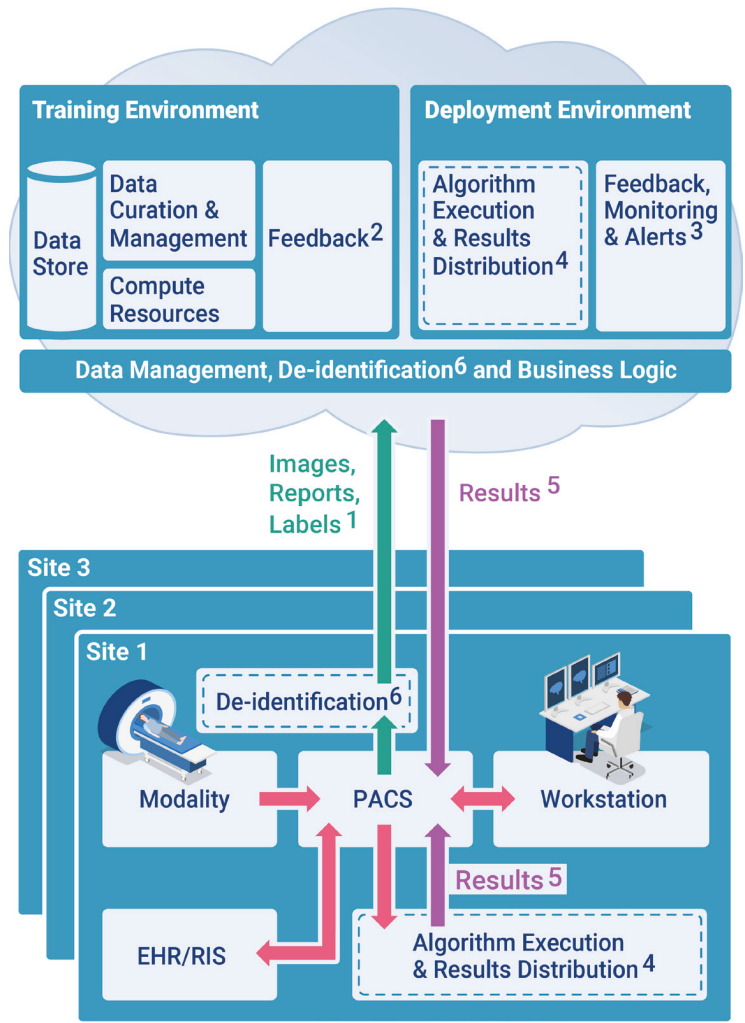

**Fig. 3.** A Centralized, Cloud Based Infrastructure for Feedback and Monitoring of AI/ML Development and Clinical Deployment.

[1] Sites contributing training data and/or running approved AI algorithms send images and related data (e.g. orders), interactive reports and labels to the cloud.

[2,3] Developed (or existing) algorithms are tested for accuracy by the feedback system, which compares labels from interactive reports to the output of the algorithm, without burdening the radiologist. When training algorithms, discrepancies and their associated data are used to iteratively improve the algorithms. For approved algorithms, the monitoring system warns relevant sites against unacceptable degradation.

[5] Results generated by approved algorithms (possibly in the form of a complete radiology report) are automatically presented to radiologists with each relevant study during the reading workflow.

[4,6] De-identification and Algorithm Execution and Result Distribution can be performed either locally, centrally, or both.

# 3. DISCUSSION

## 3.1. Value of the Approach

We believe that information readily captured from radiologists during routine clinical workflow is valuable and should not be squandered, even if retrospective labeling and localization becomes tractable or scalable. Until now, the report has been an underutilized source of re-usable data. The core gaps in many clinical radiology systems are the absence of structured data and the lack of linkage of images to the report. Both are needed to provide a source of well-characterized structured information to serve as a source of definitive labels. An interactive reporting system that combines image viewing, hyperlinking and report dictation with NLP [44] can close those gaps.

We describe a cloud-based infrastructure that collects images and labels from participating sites. It uses existing reporting methodologies of voice dictation, voice commands, NLP, and hyperlinks tightly integrated into the PACS reporting workflow. Hyperlinking causes little interruption to the routine radiologist dictation work, but results in a better report. The infrastructure allows algorithm developers to develop, tune and deploy their AI/ML models. Once deployed, the infrastructure continues to collect feedback and to monitor the algorithms' performance against real radiologist reports using the same collection strategies.

The value of this approach lies in the establishment of an AI/ML quality improvement infrastructure, leveraging labels acquired in routine reporting across multiple sites. This greatly simplifies the task of satisfying the requirements of users and regulators for safe and efficacious products. It allows for automatic deployment and ongoing enhancement of algorithms and monitoring of their performance. It provides larger, more complete cohorts for development of new algorithms, which reside in a centralized cloud repository. Gathering a vast pool of routine radiology data may also enable identification of rare and subtle cases, which might not otherwise be detected, collected and shared.

Radiologists are skeptical about AI, and the knowledge that they are contributing data to improve the algorithms that they use routinely may also serve to empower them [8], especially if they see regular performance improvements. The importance of user engagement in the AI process is well recognized [60].

This approach is applicable to multiple specialties and any form of imaging, not just radiology, since other specialists provide feedback as part of their viewing, analysis and documentation of findings. Findings may be in the form of a dedicated report (such as in histopathology [61]), or part of the clinical record in the EHR (such as in dermatology or ophthalmology), either of which may incorporate interactive and multimedia features.

Leveraging the advantages of the cloud is important. Most currently deployed PACS have a significant on-premise component for image and database storage, though there is movement to image sharing over the Internet [62] and eventually purely cloud-based PACS [63]. At some A Multisite, Report-Based, Centralized Infrastructure for Feedback and Monitoring of Radiology AI/ML Development and Clinical Deploymentpoint, the centralization of data for AI/ML will change from "site to cloud" transfer to "cloud to cloud" transfer. Since most sites already require an off-site backup mechanism for business continuity and disaster recovery [64], a cloud-based mirror of the image and label data is often already implemented and could be leveraged [7, 65]

## 3.2. Legal, Ethical, Moral and Business Considerations

Our proposed approach involves strictly voluntary participation. The technical design requires neither the engagement of every radiologist nor every site. It is independent of considerations of legal, ethical and moral aspects of secondary re-use of data, and can be deployed consistently with different interpretations of such factors. We recognize it is vital for every contributor to balance the complex aspects of stewardship, ownership, privacy, commercial interest, exclusivity, open access and public good [66, 67]. The proposed solution adheres to ethical principles regarding selling data. A recent article suggested that it is reasonable for corporate entities to profit from AI algorithms developed from clinical data, though unethical for them to profit from the data themselves [67]. Any approach needs to be robust to inevitable changes in acceptable data practices over time [68]. Compliance with requirements for data sharing and privacy principles, including sharing across borders, is essential [69, 70].

We believe that these issues are independent of the technical solution proposed. Appropriate safeguards will permit optimizing the utility of, and maximizing the re-use of, routine clinically acquired data in a beneficial and ethically sound manner.

## 3.3. Technical Considerations

We propose the use of tightly coupled and well-integrated systems that combine reporting with labeling, and in which user experience is seamless. A single supplier can provide a unified solution, or separate image viewing and reporting systems from multiple suppliers can be integrated by customizations to achieve the necessary functionality.

There are existing standards for the encoding of image selection information (DICOM KOS) and location (DICOM SEG, RTSS, SR) and labels and measurements (DICOM SR), and their use is important for the record to be interoperable and shareable. There are as yet no standards for orchestrating the interaction between the viewing and reporting systems at the granularity required for strong labeling. Accordingly, a single system that combines both viewing and reporting, implementing the necessary elements of localizing abnormalities through hyperlinking or other mechanisms, and NLP [44], is well positioned to provide relatively strong labels without additional effort by the user and without extra cost.

The human-driven reporting process often involves consolidation of information from multiple sources, including pre-population with procedure information [71], machine or technologist acquired structured data, such as contrast [72], radiation dose [73] or measurements [74, 75], as well as assistance by decision support structured templates and guidelines [76, 77]. Such clinical decision support mechanisms may be leveraged for AI/ML result incorporation during radiology reporting, and may enable comparison of the machine and radiologist findings [78]. An infrastructure that provides for recording and centralizing image findings in the form of a registry [78, 79], would also enable feedback and monitoring, but would likely not address the need for localization, unless the source structured data includes links to images and finding locations, and a means to access the images.

There are additional issues with report extracted labels that must be addressed when labeling is based on hyperlinks in reports. These include the common situation in which multiple findings are present in a study, and matching image-related hyperlinks with dictated report information. Hyperlinks provide an advantage in that the localized image information is directly linked to corresponding information in the body of the report. There is also the significance of the absence of placement of a hyperlink by the radiologist; this may imply normality, or simply indicate that a finding wasn't worth the effort of hyperlinking. Hyperlinks to normal findings as a means of illustration (perhaps in response to an explicit question in the request) may also be confounding. Hyperlinks may also be implicitly biased towards more important or more subtle findings (perhaps those that particularly need pointing out). Radiologists might also mark only representative lesions, not all of them.

In addition, there is variation between experts [80], and combining multiple opinions for the same case may improve the strength of a label [81]. It is unusual for more than one radiologist to read the same case, except during peer review or in a training environment. However, we expect that the amount of data gathered from much larger numbers of individual contributors reporting many more cases may compensate for any actual or postulated relative weakness compared to what a consensus of experts dedicated to the task might obtain.

Such details will need to be addressed with respect to the strength and quality of the labels obtained, perhaps with refinements in the reporting interface, or by radiologist training. Data gathered from initial deployments will permit the improvement of the viewer, the reporting system and data gathering platform, not just the algorithms to which it is applicable.

It is tempting to include an ever more expansive set of data elements in what is gathered from the sites, particularly to increase the strength of the labels and bring the performance monitoring closer to the actual patient outcomes. For example, the radiologist's clinical report does not contain the final diagnosis, which may reside in the EHR. Though HL7 feeds from the EHR can be incorporated in our proposed solution, there are risks in receiving a comprehensive unfiltered stream [82, 83]. Our initial focus on the radiology report and the image-associated content is limited, but it is aligned with a sufficient set of use cases to be adequate as a first step. More information from related systems can be incorporated in the automated data capture as the value of additional elements for model improvement is demonstrated.

### 3.4. Comparison with Alternative Approaches

The effort required for retrospective expert review outside of the clinical routine is considerable. It has been quantified for some tasks, and there is recognition that it may not be sustainable beyond short bursts of activity and with limitations on the total effort per expert that is feasible (or affordable) [12]. Retrospective expert review also does not provide a solution for continuous performance monitoring.

Expert labeling and annotation platforms have been developed that seek to minimize the amount of effort required, and which optimize the capture of specific labels with or without localization [12, 29, 45, 84, 85]. Such systems long predate the specific use of labels for modern ML, but were applicable for mammography CAD research and teaching [86].

Dikici et al [29] describe a system that is dedicated to the purpose of acquiring feedback, but its use requires a custom-built tool and is outside the routine clinical workflow. They recognize that the use of DICOM standards allowing for interoperability of annotations mitigates this concern somewhat. They define stages of algorithm maturity in terms of distinct levels of research, production and feedback. Their level of feedback maturity can be achieved using existing integrated off-the-shelf interactive multimedia reporting tools. These may be scaled to a multisite level, without requiring every user or site to learn and use a specialized tool, or limiting the generalizability of the result.

Another alternative approach is crowdsourcing, rather than use of carefully selected subject matter experts. This may produce weaker labels but on a larger scale at lower cost, and is theoretically possible even for image classification and diagnostic tasks [87, 88]. Existing image data sets with a known diagnosis can be augmented with localization and measurement information by such methods [84, 85]. It remains to be seen whether a crowdsourcing approach can be used for regulated devices. Crowdsourcing may produce labels for training and testing, but does not address the need for feedback and monitoring.

Structured reports (consistently organized reports with a standard outline, itemized sections and language) have long been touted as advantageous for data extraction. Yet radiologists have proven recalcitrant to their adoption on a large scale, particularly if additional effort is required during authoring [89]. They are slowly gaining favor and have renewed potential due to the demand for AI labels, particularly when localization is not required [90]. Despite the existence of standards for encoding structured reports in an interchangeable form, and recognition of their utility for AI [91], vendor adoption is poor. Notwithstanding this reluctance, reporting systems that support structured reporting should also support image labeling for AI. Relatively strong labels could be captured without burdening the reporting radiologist, particularly if localization information is also recorded.

Without a change in radiologists' incentives, it seems that a more pragmatic approach than conventional structured data entry is required, one that leverages what can be captured through existing practice, to mitigate the risk of a "productivity nightmare" [89]. The use of disease-specific templates during conventional dictation with speech recognition coupled with NLP would facilitate structured data entry, as can the use of assistive technology during dictation [92]. Scalability of NLP for global use in a multitude of languages remains challenging.

Others have proposed the centralization of data with an "honest broker", as might be provided by professional societies [93], trade organizations, industrial consortia or other third parties, whether for profit or not. We suggest it is more practical to leverage an existing deployed PACS infrastructure, and make use of the system supplier to manage, host and make use of the centralized information. Our approach obviates the need to develop, interface, deploy and test additional tools at the source sites, which may be accompanied by non-trivial capital or operational costs. It also allows existing organizational and business relationships between site and provider to be leveraged and reused, without the need for cooperation with an additional external entity that may have a potentially different political or strategic agenda.

Various flavors of federated, distributed or "no peek" learning mechanisms have been proposed that leave the images and labels at each site and instead involve redistributing code or model weights [94, 95]. These have been demonstrated for radiology applications [96-98]. It remains to be seen how successful or practical such approaches will be on a large scale. Standardizing packaging of algorithms or model weights of a family of algorithms, for interchange between different platforms at different sites, is elusive.

The federated approaches require each site to use standardized labels and interchange formats, provide data warehousing and curation services and operate dedicated computational resources. This is challenging enough for major academic centers but likely to be impractical for small community hospitals and imaging centers from whom data is essential to achieve generalizability and avoid bias. Each site needs to deploy its own local feedback and monitoring solution if it is unwilling to centralize images and labels. A combination of federated learning approaches for training, and centralized feedback and monitoring, could coexist.

In federated schemes, the distribution of incentives such as credit and revenue may also pose business model challenges.

## 4. CONCLUSIONS

As the tools of artificial intelligence play an increasing role in the medical and business practice of radiology, consideration must be given to means of collecting and curating the data through which algorithms are trained. Mechanisms must be implemented to allow for the ongoing, site-specific evaluation of deployed algorithms; such surveillance will be required by regulators. Updated algorithms with enhanced performance can then be re-deployed.

We propose a cloud-based multisite infrastructure to accomplish these goals. We have defined its essential technical and operational characteristics. It includes the centralized collection of large amounts of de-identified image data from sites around the world, and the incorporation of tools to label the images seamlessly and accurately as the studies are routinely interpreted by radiologists.

The business model for AI/ML remains elusive at this nascent stage of development [99, 100]. We feel that any approach that can monitor and maximize quality, while at the same time minimizing effort and cost, should be explored.

A Multisite, Report-Based, Centralized Infrastructure for Feedback and Monitoring of Radiology AI/ML Development and Clinical Deployment**[67]** D.B. Larson, D.C. Magnus, M.P. Lungren, N.H. Shah, C.P. Langlotz, Ethics of using and sharing clinical imaging data for artificial intelligence: A proposed framework, Radiology 295(3) (2020) 675-682, https://doi.org/10.1148/radiol.2020192536.

**[68]** J.R. Geis, A. Brady, C.C. Wu, J. Spencer, E. Ranschaert, J.L. Jaremko, S.G. Langer, A.B. Kitts, J. Birch, W.F. Shields, R. van den Hoven van Genderen, E. Kotter, J.W. Gichoya, T.S. Cook, M.B. Morgan, A. Tang, N.M. Safdar, M. Kohli, Ethics of artificial intelligence in radiology: Summary of the joint European and North American multisociety statement, Insights Imaging 10(1) (2019) 101, https://doi.org/10.1186/s13244-019-0785-8.

**[69]** X. Larrucea, M. Moffie, S. Asaf, I. Santamaria, Towards a GDPR compliant way to secure European cross border healthcare industry 4.0, Computer Standards & Interfaces 69 (2020) 103408, https://doi.org/10.1016/j.csi.2019.103408.

**[70]** AMA privacy principles. 2020, 2020, http://www.ama-assn.org/system/files/2020-05/privacy-principles.pdf, (Accessed on July 17, 2020).

**[71]** C.M. Hawkins, S. Hall, J. Hardin, S. Salisbury, A.J. Towbin, Prepopulated radiology report templates: A prospective analysis of error rate and turnaround time, J Digit Imaging 25(4) (2012) 504-511, https://doi.org/10.1007/s10278-012-9455-9.

**[72]** S. Goldberg-Stein, D. Gutman, O. Kaplun, D. Wang, A. Negassa, M.H. Scheinfeld, Autopopulation of intravenous contrast type and dose in structured report templates decreases report addenda, JACR 14(5) (2017) 659-661, https://doi.org/10.1016/j.jacr.2016.10.017.

**[73]** M.D. Kovacs, M.Y. Cho, P.F. Burchett, M. Trambert, Benefits of integrated ris/pacs/reporting due to automatic population of templated reports, Curr Probl Diagn Radiol 48(1) (2019) 37-39, https://doi.org/10.1067/j.cpradiol.2017.12.002.

**[74]** N.J. Hangiandreou, S.F. Stekel, D.J. Tradup, Comprehensive clinical implementation of DICOM structured reporting across a radiology ultrasound practice: Lessons learned, JACR 14(2) (2017) 298-300, https://doi.org/10.1016/j.jacr.2016.09.046.

**[75]** M.H. Scheinfeld, O. Kaplun, N.A. Simmons, J. Sterman, S. Goldberg-Stein, Implementing a software solution across multiple ultrasound vendors to auto-fill reports with measurement values, Curr Probl Diagn Radiol 48(3) (2019) 216-219, https://doi.org/10.1067/j.cpradiol.2018.09.002.

**[76]** T.K. Alkasab, B.C. Bizzo, L.L. Berland, S. Nair, P.V. Pandharipande, H.B. Harvey, Creation of an open framework for point-of-care computer-assisted reporting and decision support tools for radiologists, JACR 14(9) (2017) 1184-1189, https://doi.org/10.1016/j.jacr.2017.04.031.

**[77]** M. Kohli, T. Alkasab, K. Wang, M.E. Heilbrun, A.E. Flanders, K. Dreyer, C.E. Kahn, Bending the artificial intelligence curve for radiology: Informatics tools from ACR and RSNA, JACR 16(10) (2019) 1464-1470, https://doi.org/10.1016/j.jacr.2019.06.009.

**[78]** M. Tilkkin, AI in radiology: Regulatory, quality, and implementation issues, http://on-demand.gputechconf.com/gtc/2018/presentation/S8994-workflow-and-regulatory-challenges-to-algorithm-implementation.pdf, (Accessed on July 17, 2020).

**[79]** FDA-funded NEST program names Data Science Institute AI use case as demonstration project, 2018, http://www.acr.org/Media-Center/ACR-News-Releases/2018/FDA-NEST-Program-Names-ACR-DSI-Use-Case-as-Demo-Project, (Accessed on July 17, 2020).

**[80]** A.P. Brady, Error and discrepancy in radiology: Inevitable or avoidable?, Insights Imaging 8(1) (2017) 171-182, https://doi.org/10.1007/s13244-016-0534-1.
19

A Multisite, Report-Based, Centralized Infrastructure for Feedback and Monitoring of Radiology AI/ML Development and Clinical Deployment

A Multisite, Report-Based, Centralized Infrastructure for Feedback and Monitoring of Radiology AI/ML Development and Clinical Deployment